\def\year{2022}\relax
\documentclass[letterpaper]{article} % DO NOT CHANGE THIS
\usepackage{aaai22}  % DO NOT CHANGE THIS
\usepackage{times}  % DO NOT CHANGE THIS
\usepackage{helvet}  % DO NOT CHANGE THIS
\usepackage{courier}  % DO NOT CHANGE THIS
\usepackage{amsmath}
\usepackage[hyphens]{url}  % DO NOT CHANGE THIS
\usepackage{graphicx} % DO NOT CHANGE THIS
\usepackage{multirow}
\usepackage{amssymb}
\usepackage{comment}
\usepackage{subfig}

\usepackage{natbib}  % DO NOT CHANGE THIS AND DO NOT ADD ANY OPTIONS TO IT
\usepackage{caption} % DO NOT CHANGE THIS AND DO NOT ADD ANY OPTIONS TO IT
\DeclareCaptionStyle{ruled}{labelfont=normalfont,labelsep=colon,strut=off} % DO NOT CHANGE THIS
\usepackage[switch,mathlines]{lineno}  %
\usepackage{algorithm}
\usepackage{algorithmic}

\usepackage{newfloat}
\usepackage{listings}
\floatstyle{ruled}
\newfloat{listing}{tb}{lst}{}
\floatname{listing}{Listing}
\title{Learning Temporally and Semantically Consistent Unpaired Video-to-video Translation Through Pseudo-Supervision From Synthetic Optical Flow}
\author {
    Kaihong Wang\textsuperscript{\rm 1}\thanks{Work done as an intern at Honda Research Institute USA, Inc.},
    Kumar Akash\textsuperscript{\rm 2},
    Teruhisa Misu\textsuperscript{\rm 2}
}
\affiliations {
    \textsuperscript{\rm 1} Boston University\\
    \textsuperscript{\rm 2} Honda Research Institute USA, Inc.\\
    kaiwkh@bu.edu, kakash@honda-ri.com, tmisu@honda-ri.com
}
\begin{document}

\maketitle

\begin{abstract}
Unpaired video-to-video translation aims to translate videos between a source and a target domain without the need of paired training data, making it more feasible for real applications. Unfortunately, the translated videos generally suffer from temporal and semantic inconsistency. To address this, many existing works adopt spatiotemporal consistency constraints incorporating temporal information based on motion estimation. However, the inaccuracies in the estimation of motion deteriorate the quality of the guidance towards spatiotemporal consistency, which leads to unstable translation. In this work, we propose a novel paradigm that regularizes the spatiotemporal consistency by synthesizing motions in input videos with the generated optical flow instead of estimating them. Therefore, the synthetic motion can be applied in the regularization paradigm to keep motions consistent across domains without the risk of errors in motion estimation. Thereafter, we utilize our unsupervised recycle and unsupervised spatial loss, guided by the pseudo-supervision provided by the synthetic optical flow, to accurately enforce spatiotemporal consistency in both domains. Experiments show that our method is versatile in various scenarios and achieves state-of-the-art performance in generating temporally and semantically consistent videos. Code is available at: \textit{https://github.com/wangkaihong/Unsup\_Recycle\_GAN/}.
\end{abstract}

%%%%%%%%% BODY TEXT
\section{Introduction}

Generative Adversarial Network~\cite{GoodfellowPMXWOCB14}  has shown its versatility in various computer vision applications, especially in cross-domain image-to-image translation~\cite{IsolaZZE17} tasks. However, such tasks normally require a huge amount of pixel-level annotation in datasets and generating such dataset with aligned image pairs is not only expensive but sometimes also unnecessary when expected output is not uniquely paired with input. Consequently, unpaired image-to-image translation~\cite{ZhuPIE17} addresses this task in the absence of paired images. On this basis, unpaired video-to-video translation~\cite{BashkirovaUS18,BansalMRS18} is further introduced to translate a source domain video to its target domain counterpart. 

\begin{figure}[t!]
\centering
  \includegraphics[width=.99\linewidth]{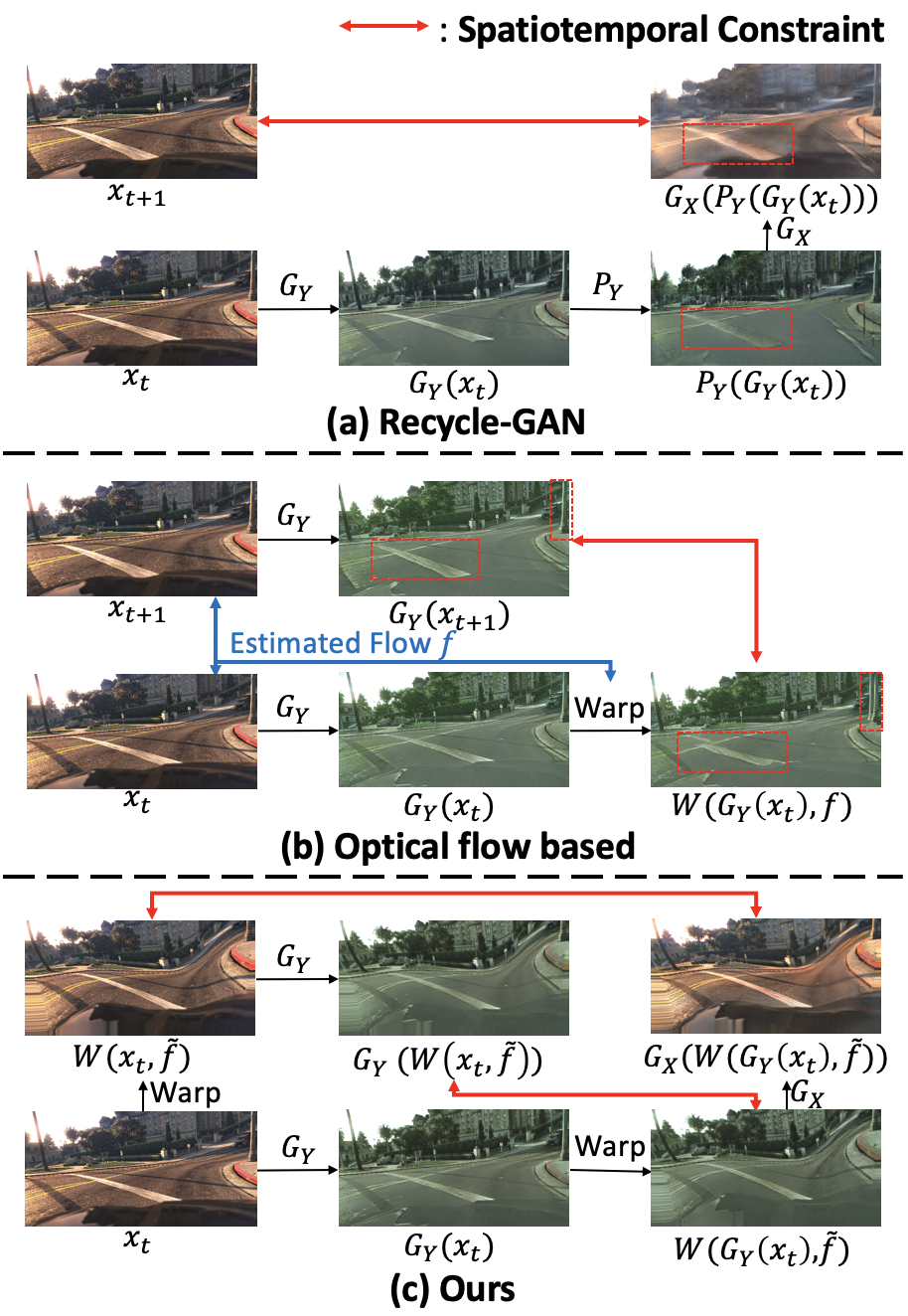}
  \caption{Comparison of spatiotemporal consistency constraints. Given an input image $x_t$ and the future frame $x_{t+1}$, spatiotemporal consistency can be enforced in various paradigms as in (a) Recycle-GAN, (b) optical flow based methods and (c) ours through the cross-domain generator $G_X$ and $G_Y$.
  Both (a) and (b) suffer from inaccuracies in motion estimation, e.g., the blurriness in $P_Y(G_Y(x_t))$ originated from the future frame predictor $P$ in (a) and mismatched regions highlighted in red rectangles originated from the warping operation $W$ with the estimated optical flow $f$ in (b), which lead to inaccurate guidance towards spatiotemporal consistency.
  In comparison, our method avoids such errors in motion estimation by using synthetic optical flow to ensure accurate regularization of spatiotemporal consistency.
  }
\label{fig:fig1}
\end{figure}

A major challenge in extending synthesis from image-level to video-level is temporal inconsistency; it originates from the subtle difference in temporally adjacent frames, which is further magnified in the unstable translation process. 
As a result, translating every frame independently leads to flickering effects even if every single frame is correctly translated to the target domain from an image translation perspective. 
On the other hand, semantic inconsistency is another common problem highly correlated with temporal inconsistency, which refers to the changes in the semantic labels during the unstable translation process.

To address these problems, Recycle-GAN~\cite{BansalMRS18} introduces spatiotemporal consistency constraint that utilizes temporal information in videos to regularize the temporal consistency in the translation, and is widely adopted by subsequent works.
Practically, the temporal information is obtained from the estimated motions between temporally adjacent frames to synchronize motions before and after the translation, thereby enforcing cyclic spatiotemporal consistency.
Meanwhile, semantic consistency is also improved when temporally contextual factors are incorporated as additional spatial smoothing.
As a result, generators are guided to perform temporally and semantically consistent video translation at the same time.

Intuitively, the effect of regularization of the spatiotemporal consistency greatly relies on the accuracy of motion estimation. 
However, most existing works fail to accurately estimate motion and therefore weaken the benefits of the regularization.
For instance, the pixel-level temporal predictor $P$ in Recycle-GAN~\cite{BansalMRS18}, as illustrated in Figure~\ref{fig:fig1}(a), and the optical flow estimators in~\cite{ChenPYTM19,ParkWKCK19}, as illustrated in Figure~\ref{fig:fig1}(b), are very likely to make inaccurate estimations of the motions due to capacity limitation of the models or domain gap problems. 
This impairs the consistency of motions across domains, and the enforcement of the spatiotemporal consistency based on the inaccurately estimated motions will inadvertently be included as the target of optimization of the generators for the translation model. 
Therefore, generators will be misled when they are forced to fix the mistakes made by motion estimation in the paradigm and synthesize consistent images given inconsistent motions. As a result, the temporal and semantic stability of translated videos is undermined.

In this paper, we propose a novel paradigm that learns temporally and semantically consistent unpaired video-to-video translation through spatiotemporal consistency regularization via pseudo-supervision from synthetic optical flow. Specifically, given a static frame from a source video, we synthesize a random optical flow, instead of estimating one, to simulate its temporal motion and synthesize the future frame through warping.
Similarly, for the target domain counterpart translated from the source frame, we simulate its future frame with the same synthetic optical flow. After that, spatiotemporal consistency regularization is enforced between the pair of future frames with the consistent motion across domains without risk of incorrect motion estimation; the synthetic optical flow acts as a flawless motion representation for both source and target domains.
Note that the quality of our pseudo-supervision does not rely on the reality of the synthetic optical flow that can simulate true motions in the domain. Instead, it focuses on ensuring the accuracy of the regularization by keeping motions across domains consistent while enforcing spatiotemporal consistency.
Besides the improvements in accuracy, our method eliminates the need for motion estimation and thereby achieves better computational efficiency as compared with the existing methods.

To construct a more reliable and efficient spatiotemporal consistency regularization paradigm, we introduce two unsupervised losses: unsupervised recycle loss and unsupervised spatial loss. Given a pair of synthetic future frames in the source and target domain generated from one input static frame, our unsupervised recycle loss enforces spatiotemporal cyclic consistency in the source domain. Meanwhile, the unsupervised spatial loss regularizes spatiotemporal consistency in the target domain.
To summarize, our main contributions in this work include:
\begin{enumerate}
  \item
We propose a novel paradigm that learns from the reliable guidance toward spatiotemporal consistency provided by the pseudo-supervision from synthetic optical flow and thereby mitigating the inaccuracies and computational costs originating from the motion estimation process.
  \item
We introduce the unsupervised recycle loss and the unsupervised spatial loss to conduct more accurate and efficient spatiotemporal consistency regularization.
  \item
We validate our method through extensive experiments and demonstrate that our method achieves state-of-the-art performance in generating temporally and semantically consistent videos.
\end{enumerate}

%------------------------------------------------------------------------
\section{Related Works}

%-------------------------------------------------------------------------
\subsection{Image-to-image \& Video-to-video Translation}
Inspired by recent progress in generative adversarial networks, several early image-to-image translation methods~\cite{DentonCSF15,RadfordMC15,ZhuKSE16,ZhaoML17,IsolaZZE17,ShettyFS18,ChoiCKH0C18,ZhuZPDEWS17} have been proposed to tackle problems in various applications such as image generating and editing. However, all these methods require costly input-output image pairs. 
Many methods~\cite{LiuT16,TaigmanPW17,ShrivastavaPTSW17} have also been proposed to address the unpaired version of the problem. 
Independently, cycle-consistency constraint has been introduced~\cite{ZhuPIE17,LiuBK17,KimCKLK17} that significantly improves the quality of translation by forcing cyclic consistency of an image before and after translations and inspires many following works~\cite{YiZTG17,HoffmanTPZISED18,HuangLBK18,YangXW18,LiuHMKALK19,BaekCUYS20}.

On the basis of image-to-image translation, the video-to-video translation problem is firstly explored in vid2vid~\cite{Wang0ZYTKC18}. %and focuses on improving the temporal consistency of resulting videos. 
The quality of synthesized videos is further optimized in~\cite{WeiZFS18,MallyaWS020} by incorporating long-term information.
Few-shot vid2vid~\cite{Wang0TLCK19} alleviates the dependency of the model on a huge dataset and enhances the generalization capability of the model. 
Similar to its image counterpart, these methods also require paired annotated videos, which is expensive to acquire.
Therefore, to address unpaired video-to-video translation, a spatiotemporal 3D translator is proposed in~\cite{BashkirovaUS18} on the basis of existing image-to-image translation networks.
To regularize spatiotemporal consistency, Recycle-GAN~\cite{BansalMRS18} integrates temporal factors into its Cycle-GAN-based model by predicting future frames.
Mocycle-GAN~\cite{ChenPYTM19} and STC-V2V~\cite{ParkWKCK19} adopt optical flow to improve the accuracy of their spatiotemporal consistency constraints. 
To further improve the coherence of the generated video, a spatiotemporal discriminator unit, as well as a self-supervised ping-pong loss, is proposed in TecoGAN~\cite{ChuXMLT20} for long-term temporal consistency regularization.
UVIT~\cite{LiuGRT21} applies an RNN-based translator as well as a self-supervised video interpolation loss to incorporate temporal information and generates videos in a recurrent manner. 
Another branch of works~\cite{BonneelTSSPP15,LaiHuWaShYuYa18,LeiXiCh20} seek to improve the temporal consistency of videos in a post-processing manner that is independent from the translating process, which can limit the quality of the improvement.

Although spatiotemporal consistency constraints in these methods improve the quality of the generated videos, these constraints rely on some form of motion estimation that is not perfect. As a result, errors in motion estimation undermine the spatiotemporal consistency constraints and thereby deteriorates the quality of the generated videos. 
To investigate the effect of the accuracy in the estimation of motions on the quality of spatiotemporal consistency constraint, we conducted a pilot experiment on STC-V2V~\cite{ParkWKCK19} with different optical flow estimating methods and calculate the warping error. The definition of the warping error will be discussed in later sections.
Besides the original method that adopts Flownet2~\cite{IlgMSKDB17} to estimate flow between two temporally adjacent frames, we also ran an experiment using the same model but with intentionally reversed order of the input frames to simulate a less accurate estimation, and another experiment with a more advanced flow estimator RAFT~\cite{TeedD20}. 
\begin{table}[t]
\centering
\begin{tabular}{c | c | c} 
 \hline
 Optical flow estimator  & EPE ($\downarrow$) & Warping Error ($\downarrow$)\\ 
 \hline\hline
 Reversed Flownet2 & - &  0.050356\\ 
 Flownet2 & 2.02 & 0.039703 \\
 RAFT & 1.43 & 0.037891 \\
 \hline
 Oracle & - & 0.029555 \\
 \hline
\end{tabular}
\caption{Pilot experiments with different optical flow estimation methods. End-point error (EPE) represents the error of the flow estimator on the train set of Sintel, indicating the error rate of the optical flow estimator, while warping error evaluates the temporal stability of the generated videos. Oracle represents the warping error in the source video.}
\label{table:pilot}
\end{table}
As presented in the results in Table~\ref{table:pilot}, the improvement of the optical flow estimating method contributes to the decrease in the warping error, which indicates the enhancement of the temporal stability of the resulting videos.   
Therefore, we conclude that regularizing the spatiotemporal consistency with the absence of accurate motion estimation limits the performance of the generator to conduct stable video translation.

Compared with existing unpaired video-to-video translation methods, our work focuses on providing more accurate and efficient spatiotemporal consistency regularization through the pseudo-supervision from synthetic optical flow instead of estimated motions between frames.

%-------------------------------------------------------------------------
\subsection{Video Style Transfer}
Different from unpaired video-to-video translation, video style transfer methods are mainly built on the basis of existing non-adversarial image style transfer and focus on the improvement of temporal consistency of the resulting videos given content videos and style images. Efficient feed-forward networks for fast video style transfer with different variants of temporal constraints~\cite{RuderDB16,
HuangWLMJZLL17,GuptaJAF17,ChenLYYH17,GaoGZY18,ChenZWSXX20} are adopted to optimize temporal consistency. 
However, all these methods either require ground truth optical flow from the dataset or suffer from the incorrect estimation of optical flow, which limits their performance. To address these problems, 
CompoundVST~\cite{WangXZWL20} proposes a single-frame based video style transfer method that adopts compound regularization to learns temporal consistency through synthetic optical flow
while LinearVST~\cite{LiLK019} and MCCNet~\cite{DengTDHMX21} seek to enhance the learning of arbitrary style transfer by modifying the architecture of models.

Noticing the drawbacks of existing spatiotemporal constraints and inspired by the recent progress in CompoundVST~\cite{WangXZWL20}, we integrate the pseudo-supervision from synthetic optical flow into our Cycle-GAN-based paradigm. Note that, compared to CompoundVST, our work focuses on the cross-domain unpaired video-to-video translation problem and not just video style transfer. 
Also, we propose a novel regularization paradigm that more effectively enforces spatiotemporal consistency in both domains and therefore facilitates not only temporally stable but also semantically consistent translation. Recent methods~\cite{EilertsenMaUn19,ZhangLiYoFu21} have also focused on enhancing video quality in different specific tasks through the regularization of consistency across frames based on motion. Nevertheless, we note that our method addresses the more general problem of unpaired video-to-video translation.

 \begin{figure}[t!]
\centering
   \includegraphics[width=1\linewidth]{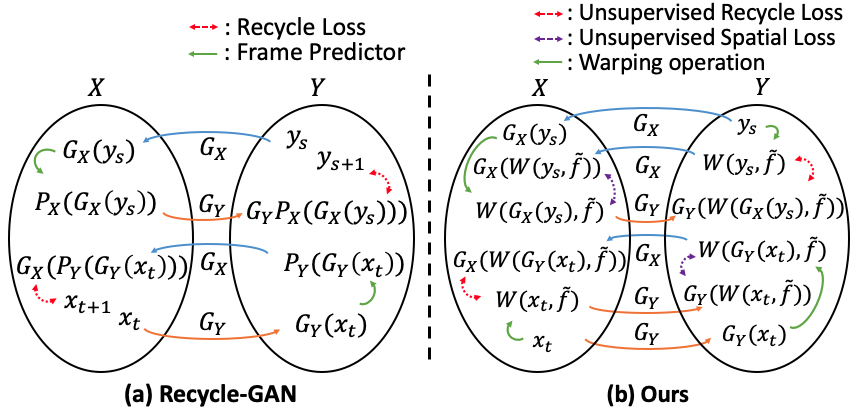}
   \caption{Illustration of the comparison between existing consistency constraint methods and ours. 
   A pair of input images $x_t$ and $y_s$ are translated through the cross-domain generators $G_X$ and $G_Y$ to the other domain.
   $P_X$ and $P_Y$ are the future frame predictor in the source and target domain, while $W(x,\Tilde{f})$ represents the warping operation on the image $x$ with the synthetic optical flow $\Tilde{f}$. (a) Recycle-GAN: Temporal information is incorporated into the recycle loss by predicting future frames in the target domain through the frame predictor $P$ and the consistency is regularized between the real future frame and the reconstructed estimated future frame in the original domain. 
   (b) Ours: Our paradigm replaces the future frame predictor with a warping operation given a synthetic optical flow $\Tilde{f}$ to simulate future frames. With the ensured accuracy in the motion simulation, we regularize the consistency of the pair of future frames in both domains with our unsupervised recycle and spatial loss.
}
\label{fig:fig2}
\end{figure}

\section{Method}

In this work, we propose a paradigm that tackles the temporal and semantic inconsistency problem in unpaired video-to-video translation tasks. Given a source video $x_{1:T}=\{x_t\}_{t=1}^T$ with $T$ frames from the source domain~$X$, and a target video $y_{1:S}=\{y_s\}_{s=1}^S$ with $S$ frames from the target domain~$Y$, we aim to learn a mapping $G_Y : X \rightarrow Y$ as well as the opposite mapping $G_X : Y \rightarrow X$ implemented on two generators without any paired videos or frames as input. With the learnt generators, target videos are synthesized as $\hat{x}_{1:T}=\{\hat{x}_t\}_{t=1}^T$ where $\hat{x}_t=G_Y(x_t)$, and $\hat{y}_{1:S}=\{\hat{y}_s\}_{s=1}^S$ where $\hat{y}_s=G_X(y_s)$.

Compared with most existing methods that translate videos in a recurrent manner, our paradigm achieves stable video-to-video synthesis through a pair of translators taking only a single frame as input at a time.
Our method consists of two objectives: an adversarial loss that guides the translation of an image to the other domain and a pair of unsupervised losses that keep videos temporally and semantically consistent across translation.
The illustration of our paradigm compared with Recycle-GAN is shown in Fig.~\ref{fig:fig2}.

\subsection{Adversarial Loss}
To translate videos to the other domain, we train the generators and discriminators adversarially in a zero-sum game as in~\cite{GoodfellowPMXWOCB14}. Practically, we optimize our discriminators $D_X/D_Y$ so that they can correctly distinguish synthetic images $\hat{x}_t/\hat{y}_s$ from real images $x_t/y_s$. At the same time, generators are trained to synthesize realistic images that can fool the discriminator. Therefore, the adversarial loss $\mathcal{L}_{adv}$ is given by

\begin{equation} \label{loss_adv}
\begin{split}
\min_G\max_D \mathcal{L}_{adv}=\sum_{s}\log D_Y(y_s)+\sum_{t}\log (1-D_Y(\hat{y}_s)) \\
 +\sum_{t}\log D_X(x_t)+\sum_{s}\log (1-D_X(\hat{x}_t)).
\end{split}
\end{equation}

\subsection{Unsupervised Losses}

Given a source frame $x_t$ as the $t$-th frame from source video~$x_{1:T}$, we simulate the synthetic future frame $\Tilde{x}_{t+1}$ by warping the current frame with a synthetic optical flow $\Tilde{f}$ as $\Tilde{x}_{t+1} = W(x_t,\Tilde{f}) = F(x_t,\Tilde{f}) + \Delta$ as in CompoundVST~\cite{WangXZWL20}.
Here $F$ represents the warping operation with the synthetic optical flow and $\Delta$ represents pixel-level random Gaussian noise. Refer to Appendix A for more details regarding the generation of the synthetic optical flow.

\textbf{Unsupervised Recycle Loss:} 
Given an original frame $x_t/y_s$ and its translated frame $\hat{x}_t/\hat{y}_s$, we simulate their synthetic following frame $\Tilde{x}_{t+1}/\Tilde{y}_{s+1}$ and $W(\hat{x}_t,\Tilde{f})/W(\hat{y}_s,\Tilde{f})$ through a warping operation with an identical synthetic optical flow $\Tilde{f}$. 
After that, we map $W(\hat{x}_t,\Tilde{f})/W(\hat{y}_s,\Tilde{f})$ back to the original domain as in Recycle-GAN and enforce its consistency with $\Tilde{x}_{t+1}/\Tilde{y}_{s+1}$ using unsupervised recycle loss $\mathcal{L}_{ur}$ given by
\begin{equation} 
\begin{split}
\mathcal{L}_{ur}=\sum_{t}{||\Tilde{x}_{t+1} - G_X(W(\hat{x}_t,\Tilde{f}))||_1} \\
+ \sum_{s}{||\Tilde{y}_{s+1} - G_Y(W(\hat{y}_s,\Tilde{f}))||_1}.
\end{split}\label{equ:ur}
\end{equation}
Compared with Recycle-GAN~\cite{BansalMRS18} that applies a pixel-level frame predictor to estimate the future frame, our paradigm uses the synthetic optical flow that ensures the consistency of the motion across domains. Consequently, the error from motion estimation becomes irrelevant to the translation process resulting in guaranteed accuracy of spatiotemporal consistency guidance. 

\textbf{Unsupervised Spatial Loss:} In parallel with $\mathcal{L}_{ur}$, our method also applies an unsupervised spatial loss $\mathcal{L}_{us}$ that enforces the consistency between $\Tilde{x}_{t+1}/\Tilde{y}_{s+1}$ and $W(\hat{x}_t,\Tilde{f})/$W($\hat{y}_s,\Tilde{f}$) in the opposite direction(see Equation \eqref{equ:us}). Practically, $\Tilde{x}_{t+1}/\Tilde{y}_{s+1}$ is mapped to the other domain to be consistent with $W(\hat{x}_t,\Tilde{f})/$W($\hat{y}_s,\Tilde{f}$) so that the warping and the translation is commutative to each other.
\begin{equation}
\begin{split}
\mathcal{L}_{us}=\sum_{t}{||G_Y(\Tilde{x}_{t+1}) - W(\hat{x}_t,\Tilde{f})||_1} \\
+ \sum_{s}{||G_X(\Tilde{y}_{s+1}) - W(\hat{y}_s,\Tilde{f})||_1}
\end{split}\label{equ:us}
\end{equation}
Guided by the pseudo-supervision from synthetic optical flow, our unsupervised losses include errors only from inconsistent translation processes in $G_X$ and $G_Y$, and minimizing the loss value leads to direct improvement in the quality of translation without misleading the generators.

\subsection{Full Objective}
Combining all loss functions we listed above, we present the full objective of our paradigm to optimize a single-frame based unpaired video-to-video translation system as
\begin{equation} \label{loss_all}
\begin{split}
\mathcal{L}=\mathcal{L}_{adv}+\lambda_{ur}\mathcal{L}_{ur}+\lambda_{us}\mathcal{L}_{us}.%+\lambda_{cont}\mathcal{L}_{cont}
\end{split}
\end{equation}
The weight of our unsupervised losses $\lambda_{ur}$ and $\lambda_{us}$ are set to 10 for all experiments. Note that additional losses can be added to the objective as needed for the translation task.

\begin{figure}[t!]
\centering
   \includegraphics[width=1\linewidth]{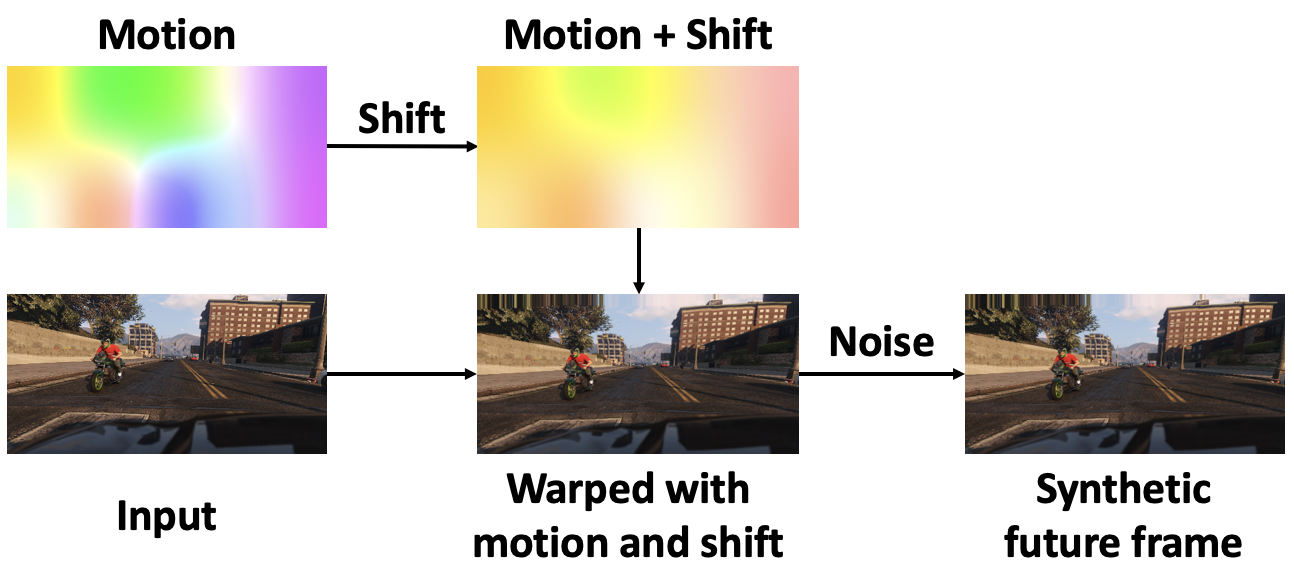}
  \caption{Synthesis of optical flow and future frame.
  }
   \label{fig:flow}
\end{figure}

\begin{figure*}[t!]
\centering
  \includegraphics[width=1\linewidth]{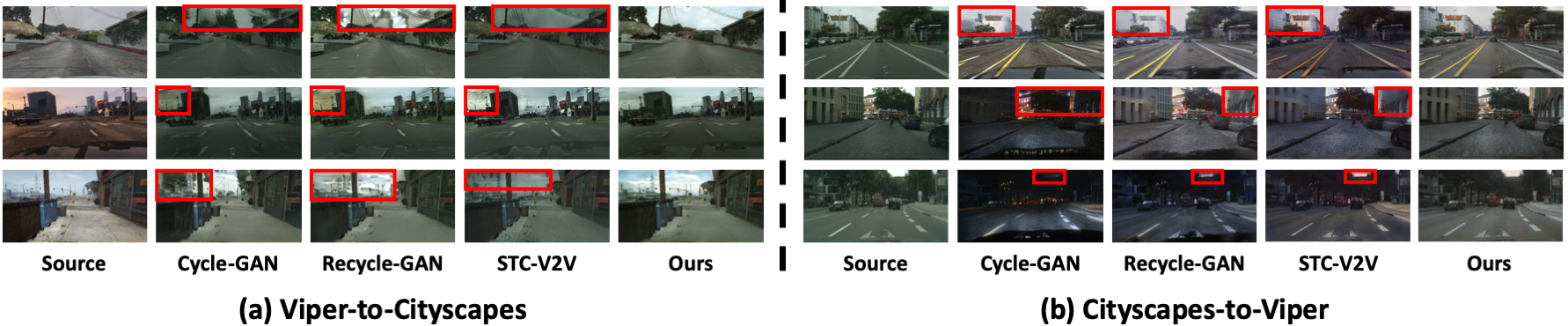}
  \caption{Comparison of existing methods with ours in the Viper-to-Cityscapes translation experiments. Based on the highlighted regions, we observe significantly fewer semantic errors. For example, in (a), the regions of the sky are more detailed and reasonable for Viper-to-Cityscapes translation. Similarly, in (b), the regions of buildings in the first and second rows as well as the regions of sky in the third row are better translated for Cityscapes-to-Viper translation.}
\label{fig:fig3}
\end{figure*}

\section{Experiments}
We evaluate and compare our method with existing methods on various aspects including temporal and semantic consistency as well as human preference. Experiments were conducted in several video-to-video translation tasks that include the intra-modality Viper-to-Cityscapes translation between driving scenes in Viper~\cite{RichterHK17} and Cityscapes~\cite{CordtsORREBFRS16} along with the inter-modality video-to-labels translation between the driving scenes and the label maps in Viper.
We also ran experiments on a series of face-to-face translation tasks and present the results in Appendix C to showcase the versatility of our method. Refer to Appendix A for further details regarding the datasets, architectures, training protocols, and other specific settings applied in the experiments.
Following our baselines~\cite{ZhuPIE17,BansalMRS18}, we set the weights for both of our losses as 10 for generality.

\textbf{Optical flow synthesis}: We follow the protocol in CompoundVST~\cite{WangXZWL20} to synthesize optical flow, which consists of motions and shifts.
For each of the $x$ and $y$ direction, the motion optical flow $\Tilde{f}_m$ is randomly generated as a grid, and each element of the grid corresponds to a $100 \times 100$ block in the input image. The grid is sampled from $\mathcal{N}(0,\,\sigma_m^{2}I)$ with $\sigma_m = 8$, and upsampled through bilinear interpolation to the original size of the input image. The shift optical flow $\Tilde{f}_s$ is generated as a constant shift sampled from $\mathcal{N}(0,\,\sigma_s^{2})$ for each of $x$ and $y$ direction with $\sigma_s = 10$. The generated synthetic optical flow $\Tilde{f} =\Tilde{f}_m+\Tilde{f}_s$ is finally smoothed using a box filter of size $100 \times 100$.
After that, a future frame can be simulated by warping an input image with the synthetic flow and adding Gaussian noise $\Delta \sim \mathcal{N}(0,\,\sigma_{n}^{2}I)$ and $\sigma_n \sim \mathcal{U}(0.01,0.02)$.
The overall process of the optical flow and future frame synthesis is illustrated in Figure~\ref{fig:flow}.

\subsection{Viper-to-Cityscapes Experiments}
In this experiment, we validate the temporal consistency and visual quality of the translated videos from our method.
Cycle-GAN~\cite{ZhuPIE17}, Recycle-GAN~\cite{BansalMRS18}, and STC-V2V~\cite{ParkWKCK19} are included as existing state-of-the-art methods for comparison. Additionally, we adopt the content loss from STC-V2V during the training of our method and set $\lambda_{cont}$ as $1$ to constraint the similarity of content in images during translation. 
Qualitative results are presented in Figure~\ref{fig:fig3}. 

\subsubsection{Temporal consistency}
First, we quantitatively evaluate the temporal consistency in the Viper-to-Cityscapes translation using warping error as the metric as presented in STC-V2V. Practically, the error $\mathcal{L}_{warp}$ is calculated as the occlusion-mask-weighted Manhattan distance between adjacent frames in videos, given ground truth optical flow $f_{t\Rightarrow t-1}$ from Viper and is given by
\begin{equation} \label{Warp_error}
\begin{split}
\mathcal{L}_{warp}=\sum_{t=2}^{K}O_{t\Rightarrow t-1}||\hat{x}_t-F(\hat{x}_{t-1},f_{t\Rightarrow t-1})||_1 .
\end{split}
\end{equation}
Here $O_{t\Rightarrow t-1}={\rm exp}(-\alpha ||x_t-F(x_{t-1},f_{t\Rightarrow t-1})||_2) $ is the occlusion mask estimated based on the difference in adjacent source frames $x_t$ and $x_{t-1}$, and $\alpha$ is set to $50$. Intuitively, lower warping error indicates a lower difference in the adjacent generated frames, implying stronger video temporal stability. We compare our method with three baselines as well as the oracle warping error in source videos, and the results are presented in Table~\ref{table:warping}. We observe that our method reduces the warping error compared with existing methods and approximates closer to the oracle level. This shows that our paradigm achieves state-of-the-art performance in producing temporally consistent videos.

\begin{table}[t]
\centering
\begin{tabular}{c | c} 
 \hline
 Method & Warping Error ($\downarrow$) \\ 
 \hline\hline
 Cycle-GAN & 0.059346 \\ 
 Recycle-GAN& 0.052738 \\
 STC-V2V & 0.039703 \\
 Ours & \textbf{0.035984} \\ 
 \hline
 Oracle(Source) & 0.029555 \\
 \hline
\end{tabular}
\caption{Comparison of warping error between our method with Cycle-GAN, Recycle-GAN, and STC-V2V as well as oracle results from the input source video for Viper-to-Cityscapes translation.}
\label{table:warping}
\end{table}

\subsubsection{User studies}
To more comprehensively evaluate the visual quality of the generated videos, we conducted user studies deployed on Amazon Mechanical Turk (AMT) and tested our method against the other baselines.
Twenty one participants were invited to review pairs of videos generated by our method versus each of the baselines in both the Viper-to-Cityscapes and the Cityscapes-to-Viper translation for the videos in the validation set of Viper and Cityscapes.
Each participant was given 15 pairs of videos consisting of comparisons between ours against the three baselines from 5 randomly sampled videos from both domains. 
In every comparison, they were asked to review a pair of videos for at least 7 seconds and select their preference in terms of consistency and quality.
We calculate the user preference in percentage according to the frequency of their selections for all three comparisons against baselines, and report the results in Table~\ref{table:user_preference}.
\begin{table}[b]
\centering
\begin{tabular}{c | c | c} 
 \hline
 Method & Forward &  Backward \\ 
 \hline\hline
 Ours / Cycle-GAN & \textbf{90.48} / 9.52 & \textbf{92.38} / 7.62 \\ 
 Ours / Recycle-GAN& \textbf{79.05} / 20.95 & \textbf{75.24} / 24.76 \\
 Ours / STC-V2V & \textbf{75.24} / 24.76 & \textbf{80.95} / 19.05\\
 \hline
\end{tabular}
\caption{Comparison of user preference between our method with others. Forward indicates user preference in percentage for the translation from Viper to Cityscapes, while backward indicates that for the translation from Cityscapes to Viper. }
\label{table:user_preference}
\end{table}
Results indicate that our method produces more visually pleasing and temporally stable videos in both directions compared with existing baselines.

\begin{figure*}[t]
\centering
   \includegraphics[width=1\linewidth]{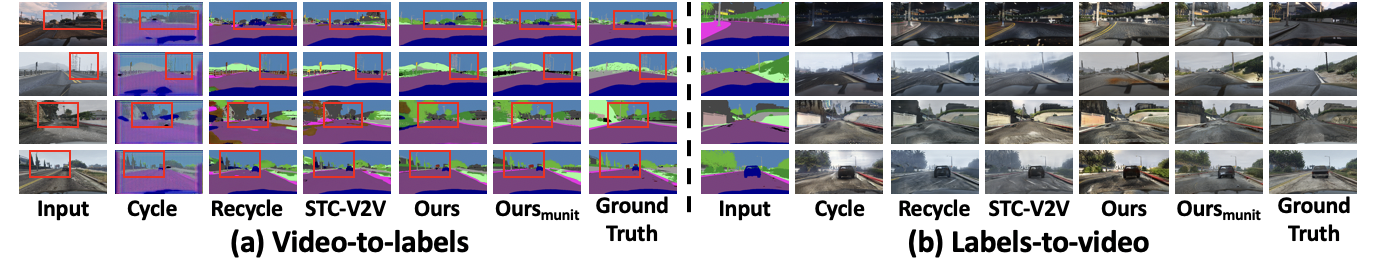}
  \caption{Comparison of existing methods with ours in the video-to-labels translation experiments.
  We can notice more accurate and detailed segmentation results for our method around the vegetation in the first, second, and fourth row; poles in the third row; and the trees in the fifth row.
  Similarly, in (b), we can also observe more realistic, natural, and detailed scenes, such as roads and sky in the third and fourth rows as well as vegetation at the first, second, and fifth rows from our results.
  }
   \label{fig:v2l}
\end{figure*}

\subsection{Video-to-Labels Experiments}
To further verify that our method performs semantically consistent video-to-video translation, we conducted the video-to-labels and labels-to-video experiments following protocols in previous works~\cite{BansalMRS18, ChenPYTM19,LiuGRT21}. In this scenario, we translate the videos to their corresponding pixel-level semantic segmentation labels in Viper on the basis of the architecture of Cycle-GAN~\cite{ZhuPIE17} and MUNIT~\cite{HuangLBK18}. Besides our unsupervised losses, we also applied the cycle loss following~\cite{ZhuPIE17,BansalMRS18} and set $\lambda_{cyc}$ as $10$ during our training process to further enhance the regularization for the sake of fair comparison.
The quality of the translation, in terms of semantic consistency, was evaluated quantitatively in both directions. 
We incorporated Cycle-GAN~\cite{ZhuPIE17}, Recycle-GAN~\cite{BansalMRS18}, and STC-V2V~\cite{ParkWKCK19} as baseline methods in the following experiments and compare their performance with ours. 
Additionally, we suppressed the regularization for the video domain in our paradigm to avoid generating vague contents from the regions in the label domain that have the same semantic labels. Further discussions for this setting are available in ablation studies and Appendix A.
Qualitative results are shown in Figure~\ref{fig:v2l}.

\begin{table}[t!]
\centering
\setlength{\tabcolsep}{1.5pt}
\fontsize{9}{10} \selectfont
\begin{tabular}{ c |c |cccccc} 
 \hline
 Criterion & method & Day & Sunset & Rain & Snow & Night & All \\ 
 \hline\hline
 \multirow{6}{*}{MP} & Cycle-GAN & 29.68 & 32.89 & 34.87 & 28.65& 17.92 & 28.29 \\ 
 &MUNIT & 39.92 & 66.54& 70.46& 64.20& \textbf{50.74} & 56.40 \\ 
 &Recycle-GAN & 46.89 & 69.77& 60.10 & 56.98 & 31.88 & 52.10  \\
 & STC-V2V & 40.98 &67.51 &62.99 & 61.86&36.07 & 52.49  \\
 &Ours & 56.31 & 76.90& 71.75& 67.86& 48.79 & 63.20 \\ 
 &Ours$_{munit}$ & \textbf{58.69} & \textbf{78.58}& \textbf{74.08}& \textbf{71.06}& 44.82 & \textbf{64.36} \\ 
 \hline
 \multirow{6}{*}{AC} & Cycle-GAN & 7.54 & 6.15 & 6.08 & 6.45 & 4.05 & 6.56  \\ 
 & MUNIT & 13.38 & 14.66& 16.76& 15.40& \textbf{11.70} & 14.36 \\ 
 & Recycle-GAN & 13.62 & 13.80 & 11.47 & 13.62 & 6.91 & 13.40  \\
 & STC-V2V  & 12.23 &13.54 & 13.34& 13.66& 6.69& 12.32  \\
 & Ours & 16.58 & 17.22&15.79 &16.14 & 8.27& 16.13 \\ 
 & Ours$_{munit}$ & \textbf{18.37} &\textbf{19.83}& \textbf{19.07} &\textbf{18.50} & 10.12& \textbf{18.34} \\ 
 \hline
 \multirow{6}{*}{mIoU} & Cycle-GAN & 4.11 & 3.66 & 3.34 & 3.56 & 1.94 & 3.76  \\ 
 & MUNIT & 7.58 & 10.17& 12.05& 11.04& \textbf{7.40} & 9.73 \\ 
 & Recycle-GAN & 9.02 & 10.90 & 8.07 & 9.60 & 3.58 & 9.53  \\
 & STC-V2V~  & 8.06 &10.61 & 9.22& 10.19& 3.97& 9.09  \\
 & Ours & 11.90 & 13.49& 11.75& 12.60& 5.14 & 12.29 \\ 
 & Ours$_{munit}$ & \textbf{13.29} & \textbf{15.64}& \textbf{13.97}& \textbf{14.17}& 6.49 & \textbf{13.71} \\ 
 \hline
\end{tabular}
\caption{Comparison of baseline methods with ours in segmentation score for video-to-label translation experiments on Viper. 
}
\label{table:v2l}
\end{table}

\begin{table}[t!]
\centering
\begin{tabular}{c | c  c  c} 
 \hline
 method  & MP & AC & mIoU \\ 
 \hline
 Recycle-GAN& 60.1 & 14.9 & 11.0  \\
 Mocycle-GAN & 64.9 & 17.7 & 13.2 \\ 
 \hline
 Improvement (\%) & 7.99  &  18.79 &  20.00 \\ 
 \hline
 Recycle-GAN  & 52.10 & 13.40 & 9.53  \\
 Ours & 63.20 & 16.13 & 12.29 \\ 
 \hline
 Improvement (\%) & \textbf{21.31} & \textbf{20.37} & \textbf{28.96} \\ 
 \hline\hline
 Recycle-GAN & 53.65 & 14.84 & 10.11 \\
 UVIT  & 65.20 & 17.59 & 13.07 \\ 
 \hline
 Improvement (\%) & 21.53  & 18.53 & 29.28  \\ 
 \hline
 Recycle-GAN  & 52.10 & 13.40 & 9.53  \\
 Ours$_{munit}$ & 64.36 & 18.34 & 13.71 \\ 
 \hline
 Improvement (\%)& \textbf{23.53}  & \textbf{36.87} & \textbf{43.86} \\ 
 \hline
\end{tabular}
\caption{Comparison of state-of-the-art methods with ours in video-to-label translation experiments.}
\label{table:SOTA_v2l}
\end{table}

\subsubsection{Video-to-Labels}
We conducted experiments of the translation from videos to labels, which is equivalent to video semantic segmentation, and evaluated the performance by comparing the generated labels with ground truth in the held-out validation set. We report the segmentation scores, including mean pixel accuracy (MP), average class accuracy (AC), and intersection over union (mIoU), for all baseline methods and ours in Table~\ref{table:v2l}. 
Additionally, we also integrated our unsupervised spatiotemporal consistency constraints into a multimodal translation model MUNIT and report the baseline MUNIT result as well as our result as Ours$_{munit}$ in the table. Further details regarding the integration are available in Appendix A.
Based on the results, we can observe the improvement of Recycle-GAN and STC-V2V over Cycle-GAN, revealing the effectiveness of the spatiotemporal consistency regularization.
Meanwhile, our method significantly outperforms all the baseline methods in three metrics under all weather conditions, which shows the benefit of learning accurate spatiotemporal consistency through pseudo-supervision from synthetic optical flow.
We further compare our method with the state-of-the-arts including Mocycle-GAN~\cite{ChenPYTM19}, a Cycle-GAN-based model learning from motion-guided spatiotemporal constraints, and UVIT~\cite{LiuGRT21}, an RNN-based method performing a multimodal translation. 
To be more specific, we take Recycle-GAN as the common baseline and compare their improvements in percentage over it with ours in Table~\ref{table:SOTA_v2l}. 
Since UVIT adopts a multimodal paradigm, we compare our method combined with MUNIT against it for the sake of fair comparison.
As presented in Table~\ref{table:SOTA_v2l}, our paradigm brings in greater improvements in all three segmentation metrics over the Recycle-GAN without relying on any extra architectures except for the basic discriminators and generators in Cycle-GAN, indicating the necessity and effectiveness of ensuring the accuracy in spatiotemporal consistency constraint using synthetic optical flow. 
Besides, it shows the generality and flexibility of our unsupervised losses that enable effective combination with other translation models to further enhance the translation quality.

\begin{table}[t]
\centering
\setlength{\tabcolsep}{1.5pt}
\fontsize{9}{10} \selectfont
\begin{tabular}{c | c | cccccc} 
 \hline
 Criterion & method & Day & Sunset & Rain & Snow & Night & All \\ 
 \hline\hline
 \multirow{6}{*}{MP} & Cycle-GAN & 26.70 & 47.25 & 40.26& 42.61 & 26.02 & 35.77  \\ 
 & MUNIT & 30.74& \textbf{70.48}& \textbf{53.41} &\textbf{59.14} &45.74 & \textbf{50.56}  \\ 
 & Recycle-GAN & 23.88 & 51.34 & 35.32& 45.45 & 28.32 & 36.35  \\
 & STC-V2V & 29.73& 52.79& 42.42& 47.91 & 37.25& 41.34  \\
 & Ours & 34.90& 66.05& 46.46 &55.49 &37.22 & 47.42  \\ 
 & Ours$_{munit}$ & \textbf{35.95}& 66.41& 47.62 &57.05 &\textbf{48.49} & 50.52  \\ 
 \hline
 \multirow{6}{*}{AC} & Cycle-GAN & 11.03 & 9.83 & 10.34 & 11.09 & 9.49 & 11.20  \\ 
 & MUNIT & 11.36 &16.31 & 14.53 & 14.17& 14.14& 14.36  \\ 
 & Recycle-GAN  & 10.94 & 12.11 & 11.64 & 11.38 & 10.99 & 11.56  \\
 & STC-V2V & 12.65 & 13.44 & 12.57& 12.33 & 13.01 & 13.10  \\
 & Ours &\textbf{14.86} &16.55 & 15.01 & 15.41& 14.14& 15.73  \\ 
 & Ours$_{munit}$ &\textbf{14.86} &\textbf{18.98} & \textbf{16.03} & \textbf{16.42}& \textbf{17.07}& \textbf{17.29}  \\ 
 \hline
 \multirow{6}{*}{mIoU} & Cycle-GAN & 5.76 & 6.70 & 5.70 & 6.53 & 4.83 & 6.36  \\ 
 & MUNIT &5.95 & 12.36& 7.92 & 9.25& 7.73& 8.66  \\ 
 & Recycle-GAN & 5.23 & 8.33& 6.22& 7.01& 5.61 & 6.67  \\
 & STC-V2V & 7.15& 9.16& 7.30& 7.59& 7.55& 8.06 \\
 & Ours &8.48 & 12.25& 8.96 & 10.36& 8.43& 10.08  \\ 
 & Ours$_{munit}$ &\textbf{8.55} & \textbf{13.75}& \textbf{9.09} & \textbf{11.07}& \textbf{10.37}& \textbf{11.07}  \\ 
 \hline
\end{tabular}
\caption{Comparison of baseline methods with ours in FCN-score for labels-to-video translation experiments on Viper. 
}
\label{table:l2v}
\end{table}

\begin{table}[t]
\centering
\begin{tabular}{c | c | c | c} 
 \hline
 method  & MP & AC & mIoU \\ 
 \hline\hline
 Recycle-GAN& 43.5 & 12.2 & 7.0  \\
 Mocycle-GAN & 47.6 & 16.0 & 10.1 \\ 
 \hline
 Improvement (\%) & 9.43 & 31.15 & 44.29 \\ 
 \hline\hline
 Recycle-GAN & 36.35 & 11.56 & 6.67 \\
 Ours & 47.42 & 15.73 & 10.08 \\ 
 \hline
 Improvement (\%) & \textbf{30.45} &  \textbf{36.07} & \textbf{51.12}  \\ 
 \hline
\end{tabular}
\caption{Comparison  of  Mocycle-GAN  with  ours  in labels-to-video translation experiments.}
\label{table:SOTA_l2v}
\end{table}

\subsubsection{Labels-to-Video}
Besides video-to-labels translation, we also compare the performance of existing methods with ours in the opposite direction that generates videos given labels. 
Following previous works~\cite{IsolaZZE17,BansalMRS18,ChenPYTM19}, we adopted FCN-score to evaluate the quality of the generated frames. 
The intuition behind this metric is that an FCN model, pretrained on the Viper dataset, should generally perform better on the videos translated from labels if they are close enough to the original videos. Therefore, higher FCN-scores indicate higher quality in the generated videos.
Table~\ref{table:l2v} compares the FCN-scores for all baseline methods with ours as well as its combination with MUNIT. We can observe from the results that with the refinement of the spatiotemporal consistency constraint, the quality of the generated images gradually improves, and our method achieves the best performance in all three metrics under all weather conditions.
Similar to the video-to-labels experiments, we also measure our improvement of FCN-score in percentage of over the common baseline, Recycle-GAN, with Mocycle-GAN and report the comparison in Table~\ref{table:SOTA_l2v}. 
The comparisons of our improvement over Recycle-GAN against current state-of-the-art methods indicate the advantage of our unsupervised paradigm. Without relying on any external architecture, our unsupervised spatiotemporal consistency learns from pseudo-supervision through synthetic optical flow and therefore achieves new state-of-the-art performance.
Additionally, further comparison with UVIT in this setting is available in Appendix C.

\begin{table}[t]
\centering
\setlength{\tabcolsep}{1.5pt}
% \fontsize{9}{10} \selectfont
\begin{tabular}{ c c c c |c c c | c c c} 
 \hline
 \multirow{2}{*}{UR}  & \multirow{2}{*}{US} & \multirow{2}{*}{VDRS} & \multirow{2}{*}{Cyc}  & \multicolumn{3}{c|}{V2L} & \multicolumn{3}{c}{L2V}\\ 
 & & & & mIoU& AC & MP & mIoU& AC & MP \\
 \hline\hline
  \checkmark & & & & 3.08 & 5.54 &23.88 & 3.69 & 7.18 &27.66 \\
  & \checkmark & & & 0.16 & 3.23 &4.86 & 4.75 & 8.55 &27.27 \\
  \checkmark & \checkmark & & & 9.89 & 13.39 &57.12 & 5.45 & 10.20 &29.50 \\
  \checkmark& \checkmark & \checkmark& & 9.53 & 12.96 &58.54 & 7.00 & 12.19 &37.93 \\
  \hline
  \checkmark& \checkmark & \checkmark& \checkmark & 12.29 & 16.13 &63.20 & 10.08 & 15.73 &47.42 \\
 \hline
\end{tabular}
\caption{Ablation studies on video-to-labels translation. UR represents unsupervised recycle loss, US represents unsupervised spatial loss, VDRS represents video domain regularization suppression and Cyc represents cycle loss.}
\label{table:ablation_v2l}
\end{table}

\begin{table}[t]
\centering
\begin{tabular}{ c  c | c } 
 \hline
 $\lambda_{UR}$  & $\lambda_{US}$  & Warping Error  \\ 
 \hline\hline
  1 & 10 & 0.040817   \\ % 0.0910/0.1232/0.5654   0.0440/0.0812/0.2572
  5 & 10 & 0.037563   \\ % 0.0899/0.1225/0.5844   0.0522/0.0951/0.2896
  10 & 1 & 0.059230   \\ % 0.0546/0.0841/0.4012   0.0734/0.1219/0.3804
  10 & 5 & \textbf{0.035434}   \\ % 0.0021/0.0338/0.0145 0.0281/0.0607/0.2451
  \hline
  10 & 10 & 0.035984  \\
 \hline
\end{tabular}
\caption{Ablation studies regarding hyper-parameters.
}
\label{table:ablation_hyp}
\end{table}

\begin{table}[ht]
\centering
% \setlength{\tabcolsep}{1.5pt}
% \fontsize{9}{10} \selectfont
\begin{tabular}{ c | c } 
 \hline
 Optical Flow Method & Warping Error  \\ 
 \hline\hline
  Estimated   & 0.048983  \\ %0.0091/0.0328/0.1579 0.0038/0.0225/0.0816
  Translation only & 0.050842  \\ % 0.0007/0.0410/0.0076 0.0322/0.0646/0.2761
  Scaling only    & 0.037798  \\ % 0.0001/0.0323/0.0041 0.0214/0.0534/0.2292
  \hline
  Ours        & \textbf{0.035984}  \\
 \hline
\end{tabular}
\caption{Ablation studies on different optical flow generation methods. }
\label{table:ablation_flow}
\end{table}

\subsection{Ablation Studies}
To verify the effectiveness of our unsupervised spatiotemporal consistency constraints and other relevant components, we ran ablation studies on the video-to-labels translation task. More ablation studies conducted on Viper-to-Cityscapes as well as to explore the effect of the synthetic optical flow are available in Appendix B.

For video-to-labels translation, we ran experiments with each of our loss functions and their meaningful combinations. Additionally, we also turned off the video domain regularization suppression to investigate its effects in video generation. The performance of these experiments was evaluated in both segmentation scores for the video-to-labels translation and FCN-scores for the labels-to-video translation. Based on the results shown in Table~\ref{table:ablation_v2l}, we can observe that each of our unsupervised losses cannot achieve semantically consistent translation independently, while their combination performs significantly better in the video-to-labels translation. On that basis, enabling video domain regularization suppression further improves the quality of the videos translated from labels, as it prevents local details suppression. Finally, the application of the cycle loss contributes another leap in the quality of the generated videos, showing the effectiveness of the cyclic spatial constraints in preserving semantic consistency. Qualitative results are presented in Appendix C for a more intuitive understanding.

To verify the effectiveness of using the proposed method to generate synthetic optical flow, we compare three optical flow generation methods with our paradigm: 1) optical flow estimated through Flownet2, 2) optical flow synthesized with only translation motions, and 3) optical flow synthesized with only scaling motions. We compare the warping error on the Viper-to-Cityscapes translation experiment. As presented in Table~\ref{table:ablation_flow}, estimated optical flow deteriorates the translation possibly due to artifacts of discontinuity as compared to our smooth synthetic flow. The artifacts combined with our strong consistency constraints undermines adverserial training and thereby the quality of translation.
On the other hand, optical flow with only low-level motions, such as translation or scaling, cannot provide rich information for temporal and semantic consistency for the training process.
In contrast, our synthetic flow ensures the quality of the optical flow with no flow discontinuity, providing stronger support to our effective regularization paradigm.

Finally, we tested hyper-parameter sensitivity in the Viper-to-Cityscapes translation with $(\lambda_{UR}, \lambda_{US})$  as $(1, 10)$, $(5, 10)$, $(10, 1)$, and $(10, 5)$. The resulting warping error are presented in Table~\ref{table:ablation_hyp}. The results show that when either of the weights is not large enough, the effectiveness of the consistency constraint will be compromised. On the other hand, when the ratio of the weights gets closer to 1, our method yields stable results, revealing the robustness of our method against hyper-parameters.

\section{Conclusion}
In this work, we propose a novel paradigm for unpaired video-to-video translation
% that addresses the temporal and semantic inconsistency problems utilizing pseudo-supervision via synthetic optical flow
using pseudo-supervision via synthetic optical flow resulting in improved temporal and semantic consistency.
Compared with typical approaches, our method simulates motions by synthesizing optical flow instead of estimating them to ensure the consistency of motions across domains and therefore guarantees the accuracy of the spatiotemporal consistency regularization in the cross-domain translation process.
With the guidance of the unsupervised recycle loss as well as the unsupervised spatial loss, our paradigm effectively facilitates stable video-to-video translation. 
We finally demonstrate the state-of-the-art performance of our method through extensive experiments.

\clearpage

\bibliography{UV2VT}

\begin{thebibliography}{51}
\providecommand{\natexlab}[1]{#1}

\bibitem[{Baek et~al.(2020)Baek, Choi, Uh, Yoo, and Shim}]{BaekCUYS20}
Baek, K.; Choi, Y.; Uh, Y.; Yoo, J.; and Shim, H. 2020.
\newblock Rethinking the Truly Unsupervised Image-to-Image Translation.
\newblock \emph{arXiv preprint arXiv:2006.06500}.

\bibitem[{Bansal et~al.(2018)Bansal, Ma, Ramanan, and Sheikh}]{BansalMRS18}
Bansal, A.; Ma, S.; Ramanan, D.; and Sheikh, Y. 2018.
\newblock Recycle-GAN: Unsupervised Video Retargeting.
\newblock In \emph{ECCV}.

\bibitem[{Bashkirova, Usman, and Saenko(2018)}]{BashkirovaUS18}
Bashkirova, D.; Usman, B.; and Saenko, K. 2018.
\newblock Unsupervised Video-to-Video Translation.
\newblock \emph{arXiv preprint arXiv:1806.03698}.

\bibitem[{Bonneel et~al.(2015)Bonneel, Tompkin, Sunkavalli, Sun, Paris, and
  Pfister}]{BonneelTSSPP15}
Bonneel, N.; Tompkin, J.; Sunkavalli, K.; Sun, D.; Paris, S.; and Pfister, H.
  2015.
\newblock Blind video temporal consistency.
\newblock \emph{{ACM} Trans. Graph.}

\bibitem[{Carreira and Zisserman(2017)}]{CarreiraZ17}
Carreira, J.; and Zisserman, A. 2017.
\newblock Quo Vadis, Action Recognition? {A} New Model and the Kinetics
  Dataset.
\newblock In \emph{CVPR}.

\bibitem[{Chen et~al.(2017)Chen, Liao, Yuan, Yu, and Hua}]{ChenLYYH17}
Chen, D.; Liao, J.; Yuan, L.; Yu, N.; and Hua, G. 2017.
\newblock Coherent Online Video Style Transfer.
\newblock In \emph{ICCV}.

\bibitem[{Chen et~al.(2020)Chen, Zhang, Wang, Shu, Xu, and Xu}]{ChenZWSXX20}
Chen, X.; Zhang, Y.; Wang, Y.; Shu, H.; Xu, C.; and Xu, C. 2020.
\newblock Optical Flow Distillation: Towards Efficient and Stable Video Style
  Transfer.
\newblock In \emph{ECCV}.

\bibitem[{Chen et~al.(2019)Chen, Pan, Yao, Tian, and Mei}]{ChenPYTM19}
Chen, Y.; Pan, Y.; Yao, T.; Tian, X.; and Mei, T. 2019.
\newblock Mocycle-GAN: Unpaired Video-to-Video Translation.
\newblock In \emph{ACM Multimedia}.

\bibitem[{Choi et~al.(2018)Choi, Choi, Kim, Ha, Kim, and Choo}]{ChoiCKH0C18}
Choi, Y.; Choi, M.; Kim, M.; Ha, J.; Kim, S.; and Choo, J. 2018.
\newblock StarGAN: Unified Generative Adversarial Networks for Multi-Domain
  Image-to-Image Translation.
\newblock In \emph{CVPR}.

\bibitem[{Chu et~al.(2020)Chu, Xie, Mayer, Leal{-}Taix{\'{e}}, and
  Thuerey}]{ChuXMLT20}
Chu, M.; Xie, Y.; Mayer, J.; Leal{-}Taix{\'{e}}, L.; and Thuerey, N. 2020.
\newblock Learning temporal coherence via self-supervision for GAN-based video
  generation.
\newblock \emph{ACM Transactions on Graphics}.

\bibitem[{Cordts et~al.(2016)Cordts, Omran, Ramos, Rehfeld, Enzweiler,
  Benenson, Franke, Roth, and Schiele}]{CordtsORREBFRS16}
Cordts, M.; Omran, M.; Ramos, S.; Rehfeld, T.; Enzweiler, M.; Benenson, R.;
  Franke, U.; Roth, S.; and Schiele, B. 2016.
\newblock The Cityscapes Dataset for Semantic Urban Scene Understanding.
\newblock In \emph{CVPR}.

\bibitem[{Deng et~al.(2021)Deng, Tang, Dong, Huang, Ma, and Xu}]{DengTDHMX21}
Deng, Y.; Tang, F.; Dong, W.; Huang, H.; Ma, C.; and Xu, C. 2021.
\newblock Arbitrary Video Style Transfer via Multi-Channel Correlation.
\newblock In \emph{AAAI}.

\bibitem[{Denton et~al.(2015)Denton, Chintala, Szlam, and Fergus}]{DentonCSF15}
Denton, E.~L.; Chintala, S.; Szlam, A.; and Fergus, R. 2015.
\newblock Deep Generative Image Models using a Laplacian Pyramid of Adversarial
  Networks.
\newblock In \emph{NIPS}.

\bibitem[{Eilertsen, Mantiuk, and Unger(2019)}]{EilertsenMaUn19}
Eilertsen, G.; Mantiuk, R.~K.; and Unger, J. 2019.
\newblock Single-Frame Regularization for Temporally Stable CNNs.
\newblock In \emph{CVPR}.

\bibitem[{Gao et~al.(2018)Gao, Gu, Zhang, and Yu}]{GaoGZY18}
Gao, C.; Gu, D.; Zhang, F.; and Yu, Y. 2018.
\newblock ReCoNet: Real-Time Coherent Video Style Transfer Network.
\newblock In \emph{ACCV}.

\bibitem[{Goodfellow et~al.(2014)Goodfellow, Pouget{-}Abadie, Mirza, Xu,
  Warde{-}Farley, Ozair, Courville, and Bengio}]{GoodfellowPMXWOCB14}
Goodfellow, I.~J.; Pouget{-}Abadie, J.; Mirza, M.; Xu, B.; Warde{-}Farley, D.;
  Ozair, S.; Courville, A.~C.; and Bengio, Y. 2014.
\newblock Generative Adversarial Nets.
\newblock In \emph{NIPS}.

\bibitem[{Gupta et~al.(2017)Gupta, Johnson, Alahi, and Fei{-}Fei}]{GuptaJAF17}
Gupta, A.; Johnson, J.; Alahi, A.; and Fei{-}Fei, L. 2017.
\newblock Characterizing and Improving Stability in Neural Style Transfer.
\newblock In \emph{ICCV}.

\bibitem[{Hoffman et~al.(2018)Hoffman, Tzeng, Park, Zhu, Isola, Saenko, Efros,
  and Darrell}]{HoffmanTPZISED18}
Hoffman, J.; Tzeng, E.; Park, T.; Zhu, J.; Isola, P.; Saenko, K.; Efros, A.~A.;
  and Darrell, T. 2018.
\newblock CyCADA: Cycle-Consistent Adversarial Domain Adaptation.
\newblock In Dy, J.~G.; and Krause, A., eds., \emph{ICML}.

\bibitem[{Huang et~al.(2017)Huang, Wang, Luo, Ma, Jiang, Zhu, Li, and
  Liu}]{HuangWLMJZLL17}
Huang, H.; Wang, H.; Luo, W.; Ma, L.; Jiang, W.; Zhu, X.; Li, Z.; and Liu, W.
  2017.
\newblock Real-Time Neural Style Transfer for Videos.
\newblock In \emph{CVPR}.

\bibitem[{Huang et~al.(2018)Huang, Liu, Belongie, and Kautz}]{HuangLBK18}
Huang, X.; Liu, M.; Belongie, S.~J.; and Kautz, J. 2018.
\newblock Multimodal Unsupervised Image-to-Image Translation.
\newblock In \emph{ECCV}.

\bibitem[{Ilg et~al.(2017)Ilg, Mayer, Saikia, Keuper, Dosovitskiy, and
  Brox}]{IlgMSKDB17}
Ilg, E.; Mayer, N.; Saikia, T.; Keuper, M.; Dosovitskiy, A.; and Brox, T. 2017.
\newblock FlowNet 2.0: Evolution of Optical Flow Estimation with Deep Networks.
\newblock In \emph{CVPR}.

\bibitem[{Isola et~al.(2017)Isola, Zhu, Zhou, and Efros}]{IsolaZZE17}
Isola, P.; Zhu, J.; Zhou, T.; and Efros, A.~A. 2017.
\newblock Image-to-Image Translation with Conditional Adversarial Networks.
\newblock In \emph{CVPR}.

\bibitem[{Kim et~al.(2017)Kim, Cha, Kim, Lee, and Kim}]{KimCKLK17}
Kim, T.; Cha, M.; Kim, H.; Lee, J.~K.; and Kim, J. 2017.
\newblock Learning to Discover Cross-Domain Relations with Generative
  Adversarial Networks.
\newblock In \emph{ICML}.

\bibitem[{Kingma and Ba(2015)}]{KingmaB14}
Kingma, D.~P.; and Ba, J. 2015.
\newblock Adam: {A} Method for Stochastic Optimization.
\newblock In \emph{ICLR}.

\bibitem[{Lai et~al.(2018)Lai, Huang, Wang, Shechtman, Yumer, and
  Yang}]{LaiHuWaShYuYa18}
Lai, W.; Huang, J.; Wang, O.; Shechtman, E.; Yumer, E.; and Yang, M. 2018.
\newblock Learning Blind Video Temporal Consistency.
\newblock In \emph{ECCV}.

\bibitem[{Lei, Xing, and Chen(2020)}]{LeiXiCh20}
Lei, C.; Xing, Y.; and Chen, Q. 2020.
\newblock Blind Video Temporal Consistency via Deep Video Prior.
\newblock In \emph{NeurIPS}.

\bibitem[{Li et~al.(2019)Li, Liu, Kautz, and Yang}]{LiLK019}
Li, X.; Liu, S.; Kautz, J.; and Yang, M. 2019.
\newblock Learning Linear Transformations for Fast Image and Video Style
  Transfer.
\newblock In \emph{CVPR}.

\bibitem[{Liu et~al.(2021)Liu, Gu, Romero, and Timofte}]{LiuGRT21}
Liu, K.; Gu, S.; Romero, A.; and Timofte, R. 2021.
\newblock Unsupervised Multimodal Video-to-Video Translation via
  Self-Supervised Learning.
\newblock In \emph{WACV}.

\bibitem[{Liu, Breuel, and Kautz(2017)}]{LiuBK17}
Liu, M.; Breuel, T.; and Kautz, J. 2017.
\newblock Unsupervised Image-to-Image Translation Networks.
\newblock In \emph{NIPS}.

\bibitem[{Liu et~al.(2019)Liu, Huang, Mallya, Karras, Aila, Lehtinen, and
  Kautz}]{LiuHMKALK19}
Liu, M.; Huang, X.; Mallya, A.; Karras, T.; Aila, T.; Lehtinen, J.; and Kautz,
  J. 2019.
\newblock Few-Shot Unsupervised Image-to-Image Translation.
\newblock In \emph{ICCV}.

\bibitem[{Liu and Tuzel(2016)}]{LiuT16}
Liu, M.; and Tuzel, O. 2016.
\newblock Coupled Generative Adversarial Networks.
\newblock In \emph{NIPS}.

\bibitem[{Mallya et~al.(2020)Mallya, Wang, Sapra, and Liu}]{MallyaWS020}
Mallya, A.; Wang, T.; Sapra, K.; and Liu, M. 2020.
\newblock World-Consistent Video-to-Video Synthesis.
\newblock In \emph{ECCV}.

\bibitem[{Park et~al.(2019)Park, Woo, Kim, Cho, and Kweon}]{ParkWKCK19}
Park, K.; Woo, S.; Kim, D.; Cho, D.; and Kweon, I.~S. 2019.
\newblock Preserving Semantic and Temporal Consistency for Unpaired
  Video-to-Video Translation.
\newblock In \emph{ACM Multimedia}.

\bibitem[{Radford, Metz, and Chintala(2016)}]{RadfordMC15}
Radford, A.; Metz, L.; and Chintala, S. 2016.
\newblock Unsupervised Representation Learning with Deep Convolutional
  Generative Adversarial Networks.
\newblock In \emph{ICLR}.

\bibitem[{Richter, Hayder, and Koltun(2017)}]{RichterHK17}
Richter, S.~R.; Hayder, Z.; and Koltun, V. 2017.
\newblock Playing for Benchmarks.
\newblock In \emph{ICCV}.

\bibitem[{Ruder, Dosovitskiy, and Brox(2016)}]{RuderDB16}
Ruder, M.; Dosovitskiy, A.; and Brox, T. 2016.
\newblock Artistic Style Transfer for Videos.
\newblock In Rosenhahn, B.; and Andres, B., eds., \emph{GCPR}.

\bibitem[{Shetty, Fritz, and Schiele(2018)}]{ShettyFS18}
Shetty, R.; Fritz, M.; and Schiele, B. 2018.
\newblock Adversarial Scene Editing: Automatic Object Removal from Weak
  Supervision.
\newblock In \emph{NIPS}.

\bibitem[{Shrivastava et~al.(2017)Shrivastava, Pfister, Tuzel, Susskind, Wang,
  and Webb}]{ShrivastavaPTSW17}
Shrivastava, A.; Pfister, T.; Tuzel, O.; Susskind, J.; Wang, W.; and Webb, R.
  2017.
\newblock Learning from Simulated and Unsupervised Images through Adversarial
  Training.
\newblock In \emph{CVPR}.

\bibitem[{Taigman, Polyak, and Wolf(2017)}]{TaigmanPW17}
Taigman, Y.; Polyak, A.; and Wolf, L. 2017.
\newblock Unsupervised Cross-Domain Image Generation.
\newblock In \emph{ICLR}.

\bibitem[{Teed and Deng(2020)}]{TeedD20}
Teed, Z.; and Deng, J. 2020.
\newblock {RAFT:} Recurrent All-Pairs Field Transforms for Optical Flow.
\newblock In \emph{ECCV}.

\bibitem[{Wang et~al.(2019)Wang, Liu, Tao, Liu, Catanzaro, and
  Kautz}]{Wang0TLCK19}
Wang, T.; Liu, M.; Tao, A.; Liu, G.; Catanzaro, B.; and Kautz, J. 2019.
\newblock Few-shot Video-to-Video Synthesis.
\newblock In \emph{NIPS}.

\bibitem[{Wang et~al.(2018)Wang, Liu, Zhu, Yakovenko, Tao, Kautz, and
  Catanzaro}]{Wang0ZYTKC18}
Wang, T.; Liu, M.; Zhu, J.; Yakovenko, N.; Tao, A.; Kautz, J.; and Catanzaro,
  B. 2018.
\newblock Video-to-Video Synthesis.
\newblock In \emph{NIPS}.

\bibitem[{Wang et~al.(2020)Wang, Xu, Zhang, Wang, and Liu}]{WangXZWL20}
Wang, W.; Xu, J.; Zhang, L.; Wang, Y.; and Liu, J. 2020.
\newblock Consistent Video Style Transfer via Compound Regularization.
\newblock In \emph{AAAI}.

\bibitem[{Wei et~al.(2018)Wei, Zhu, Feng, and Su}]{WeiZFS18}
Wei, X.; Zhu, J.; Feng, S.; and Su, H. 2018.
\newblock Video-to-Video Translation with Global Temporal Consistency.
\newblock In \emph{ACM Multimedia}.

\bibitem[{Yang, Xie, and Wang(2018)}]{YangXW18}
Yang, X.; Xie, D.; and Wang, X. 2018.
\newblock Crossing-Domain Generative Adversarial Networks for Unsupervised
  Multi-Domain Image-to-Image Translation.
\newblock In \emph{ACM Multimedia}.

\bibitem[{Yi et~al.(2017)Yi, Zhang, Tan, and Gong}]{YiZTG17}
Yi, Z.; Zhang, H.~R.; Tan, P.; and Gong, M. 2017.
\newblock DualGAN: Unsupervised Dual Learning for Image-to-Image Translation.
\newblock In \emph{ICCV}.

\bibitem[{Zhang et~al.(2021)Zhang, Li, You, and Fu}]{ZhangLiYoFu21}
Zhang, F.; Li, Y.; You, S.; and Fu, Y. 2021.
\newblock Learning Temporal Consistency for Low Light Video Enhancement From
  Single Images.
\newblock In \emph{CVPR}.

\bibitem[{Zhao, Mathieu, and LeCun(2017)}]{ZhaoML17}
Zhao, J.~J.; Mathieu, M.; and LeCun, Y. 2017.
\newblock Energy-based Generative Adversarial Networks.
\newblock In \emph{ICLR}.

\bibitem[{Zhu et~al.(2016)Zhu, Kr{\"{a}}henb{\"{u}}hl, Shechtman, and
  Efros}]{ZhuKSE16}
Zhu, J.; Kr{\"{a}}henb{\"{u}}hl, P.; Shechtman, E.; and Efros, A.~A. 2016.
\newblock Generative Visual Manipulation on the Natural Image Manifold.
\newblock In \emph{ECCV}.

\bibitem[{Zhu et~al.(2017{\natexlab{a}})Zhu, Park, Isola, and Efros}]{ZhuPIE17}
Zhu, J.; Park, T.; Isola, P.; and Efros, A.~A. 2017{\natexlab{a}}.
\newblock Unpaired Image-to-Image Translation Using Cycle-Consistent
  Adversarial Networks.
\newblock In \emph{ICCV}.

\bibitem[{Zhu et~al.(2017{\natexlab{b}})Zhu, Zhang, Pathak, Darrell, Efros,
  Wang, and Shechtman}]{ZhuZPDEWS17}
Zhu, J.; Zhang, R.; Pathak, D.; Darrell, T.; Efros, A.~A.; Wang, O.; and
  Shechtman, E. 2017{\natexlab{b}}.
\newblock Toward Multimodal Image-to-Image Translation.
\newblock In \emph{NIPS}.

\end{thebibliography}

\appendix

\clearpage

\section{Appendix A}
\subsection{Datasets}

\textbf{Viper}: Viper dataset is one of the largest synthetic urban scene video dataset generated from the game Grand Theft Auto V and consists of 77 videos with a resolution of $1920 \times 1080$ under various weather conditions, including day, sunset, rain, snow, night. These videos are also paired with several modalities such as pixel-level semantic labels and optical flow. Following previous works~\cite{BansalMRS18, ChenPYTM19}, we split the dataset into 56 videos as training set and 21 as testing set through stratified sampling based on their weather condition.

\textbf{Cityscapes}: Cityscapes dataset consists of images of real street scenes with a resolution of $2048 \times 1024$. It includes 2975 images in the training set and 500 images in the validation set. Based on this, Cityscapes sequential dataset is established by incorporating surrounding 30 frames for each image to extend them to videos.

\textbf{Face-to-face}: We adopt the publicly available face-to-face dataset curated and released by Recycle-GAN~\cite{BansalMRS18}. The included images are collected from videos of several celebrities, and their faces are cropped. 
It contains in total 11176 frames of faces from Barack Obama, 14262 frames from Donald Trump, 60764 frames from John Oliver, and 29316 frames from Stephen Colbert, respectively, in the training and test sets.

\subsection{Experiment settings}
\textbf{Network architecture}: For the sake of fair comparison, we construct our generators and discriminators following Recycle-GAN. Our generators consist of two convolution layers with a stride of 2, six residual blocks, and two upsampling convolution layers with a stride of 0.5. For the discriminators, we use the $70 \times 70$ PatchGAN from ~\cite{IsolaZZE17}. 

\textbf{Training protocol}: Experiments are conducted on an NVIDIA Tesla V100 GPU with a batch size of 1. The weight of our unsupervised losses $\lambda_{ur}$ and $\lambda_{us}$ are set to 10 for all experiments. During the training of the STC-V2V model in video-to-labels translation, we turned off its content loss. Following the settings in existing works~\cite{BansalMRS18, ParkWKCK19,ChenPYTM19}, for all Viper-to-Cityscapes experiments, the resolution of frames is set to $512 \times 256$, while for all video-to-labels and face-to-face experiments, the resolution is $256 \times 256$. 
As for the optimizer, we adopt Adam optimizer~\cite{KingmaB14} with a learning rate of 0.0002 and $\beta$ of 0.5 and 0.999. For all Viper-to-Cityscapes experiments, we train models for 2 epochs, while for all  video-to-labels and face-to-face experiments, we train models for 5 epochs.

\textbf{Video domain regularization suppression}:
In the video-to-labels experiments, the asymmetric one-to-many relationship between semantic label maps and video scenes breaks our assumption of the one-to-one relationship in our unsupervised spatiotemporal consistency constraints Therefore, the translators are very likely to generate blurry regions given label maps that have huge regions with identical labels.This is due to the labels less likely to be perturbed than video scenes by a synthetic optical flow, which will cause the consistency constraints in the video scene domain before and after the warping operation to depress the learning of local details. To address this problem, we suppress the regularization between $\Tilde{x}_{t+1}$ and $G_X(W(\hat{x}_t,f))$ in Equation~\ref{equ:ur} as well as $G_Y(\Tilde{y}_{s+1})$ and $W(\hat{y}_s,f)$ in Equation~\ref{equ:us}.

\textbf{Combination with MUNIT}:
Our spatiotemporal consistency constraints can be easily transplanted onto MUNIT~\cite{HuangLBK18} to perform multimodal video-to-video translation. For simplicity, we only depict the translation from the source to the target domain. In MUNIT, given a source domain image $x_t$ and a target domain image $y_s$, the translation from to target domain $\hat{x}_{t}$ and the reconstructed source domain image $x^{recon}_{t}$ could be computed as
\begin{equation} 
\begin{split}
\hat{x}_{t}=G_Y(E^c_X(x_t),E^s_Y(y_s)) \enspace \text{and}\\
x^{recon}_{t}=G_X(E^c_Y(\hat{x}_{t}),E^s_X(x_t)) .
\end{split}
\end{equation}
Here $G_X$ and $G_Y$ are the target and source domain generator, $E^c_X$ and $E^s_X$ are the source domain content and style feature extractor, and $E^c_Y$ and $E^s_Y$ are the target domain content and style feature extractor respectively. Thus, we can apply our unsupervised recycle loss in the source  to target translation on MUNIT with a synthetic optical flow $\Tilde{f}$ as
\begin{equation} 
\begin{split}
\mathcal{L}_{ur}=\sum_{t}{||W(x_t,\Tilde{f}) - G_X(E^c_Y(W(\hat{x}_t,\Tilde{f}))),E^s_X(x_t))||_1} .
\end{split}
\end{equation}
Similarly, the unsupervised spatial loss can be computed as
\begin{equation} 
\begin{split}
\mathcal{L}_{us}=\sum_{t}{||G_Y(E^c_X(W(x_t,\Tilde{f})), E^s_Y(y_s)) - W(\hat{x}_t,\Tilde{f})||_1} .
\end{split}
\end{equation}

\section{Appendix B}

% \begin{figure*}[t!]
% \begin{center}
%   \includegraphics[width=1\linewidth]{fig/Fig-3_v2.png}
% \end{center}
%   \caption{Comparison of existing methods with ours.}
% \label{fig:fig3}
% \end{figure*}

\subsection{More ablation studies}
\textbf{Ablation studies on Viper-to-Cityscapes}: We run experiments in the Viper-to-Cityscapes tasks with each component from our method, including our unsupervised losses, cycle loss as well as content loss from STC-V2V, and their meaningful combinations to test their influence on the temporal consistency of the resulting videos. Table~\ref{table:ablation_v2c} reports the warping errors with those configurations and shows the effectiveness of our unsupervised spatiotemporal consistency constraints in producing temporally consistent videos. 
Equipped with content loss, our method further reduces warping errors and achieves state-of-the-art performance.
On the other hand, the improvement brought by the deployment of cycle loss is much less significant, showing that cycle loss is not necessary to generate temporally consistent video.
\begin{table}[h!]
\centering
% \setlength{\tabcolsep}{1.5pt}
% \fontsize{9}{10} \selectfont
\begin{tabular}{ c c c c |c } 
 \hline
 UR  & US & Cyc & Cont  & Warping Error ($\downarrow$) \\
 \hline\hline
  \checkmark & & & & 0.075564 \\
  & \checkmark & & & 0.053536 \\
  \checkmark & \checkmark & & & 0.049261 \\
  \checkmark& \checkmark & \checkmark& & 0.047391 \\
  \checkmark& \checkmark & \checkmark& \checkmark & 0.035886 \\
 \hline
  \checkmark& \checkmark & & \checkmark & 0.035984 \\
 \hline
\end{tabular}
\caption{Ablation studies of our method on Viper-to-Cityscapes translation. UR represents unsupervised recycle loss, US represents unsupervised spatial loss, Cyc represents cycle loss and Cont represents content loss.}
\label{table:ablation_v2c}
\end{table}

% \begin{figure*}[t!]
% \begin{center}
%   \includegraphics[width=1\linewidth]{fig/Fig_face.png}
% \end{center}
%   \caption{More face-to-face results.
% }
% \label{fig:fig_more_face}
% \end{figure*}

\begin{figure*}[t!]
\centering
  \includegraphics[width=1\linewidth]{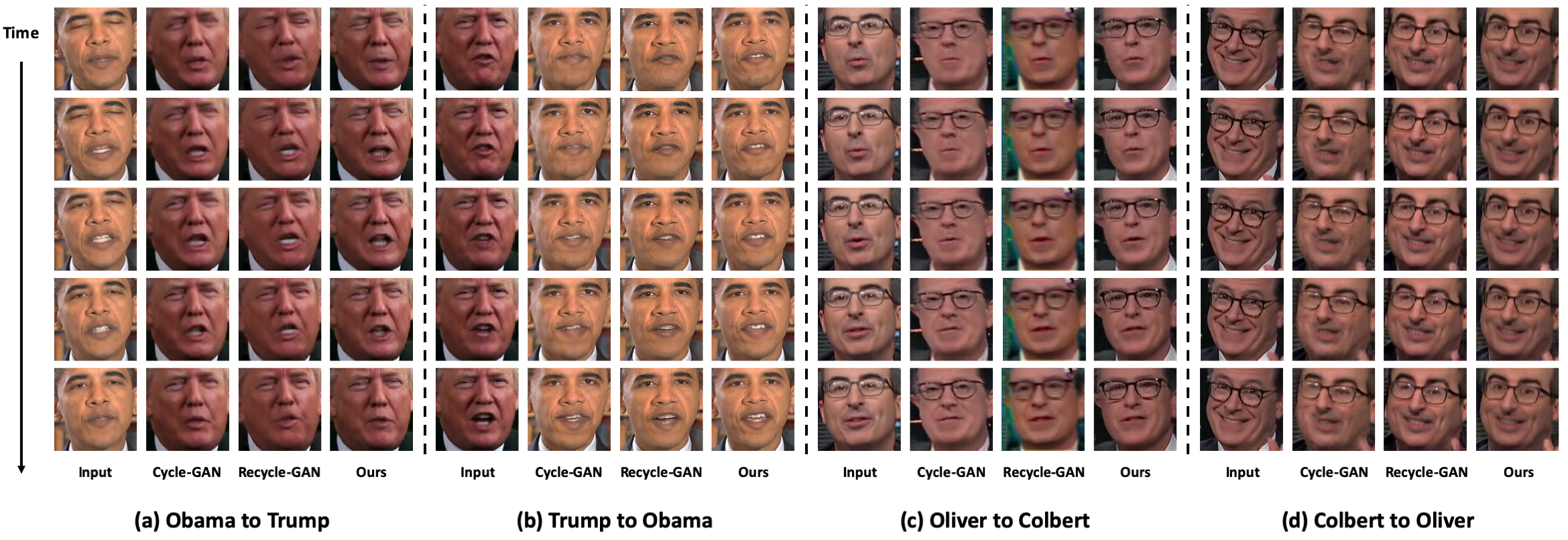}
  \caption{Illustration of our face-to-face translation results. (a) and (b) present the results in the translation between Obama and Trump, while (c) and (d) present translation between Stephen Colbert and John Oliver. 
  Clearly, Cycle-GAN miss many details in the target domain, e.g., the wrinkles between the eyebrows of Trump in (a) and the shape of the mouth of Oliver in (d). On the other hand, Recycle-GAN produces images with finer details but might introduce some local artifacts, e.g., the artifacts around the face of Colbert in the (c). Also, it keeps too many characteristics from the input instead of adapting features in the target domain. For instance, the mouth in the generated frames of Trump in (a) is still synchronized with source frames rather than moving in Trump's own way to speak.
}
\label{fig:fig_face}
\end{figure*}

\begin{figure*}[t!]
\centering
   \includegraphics[width=1\linewidth]{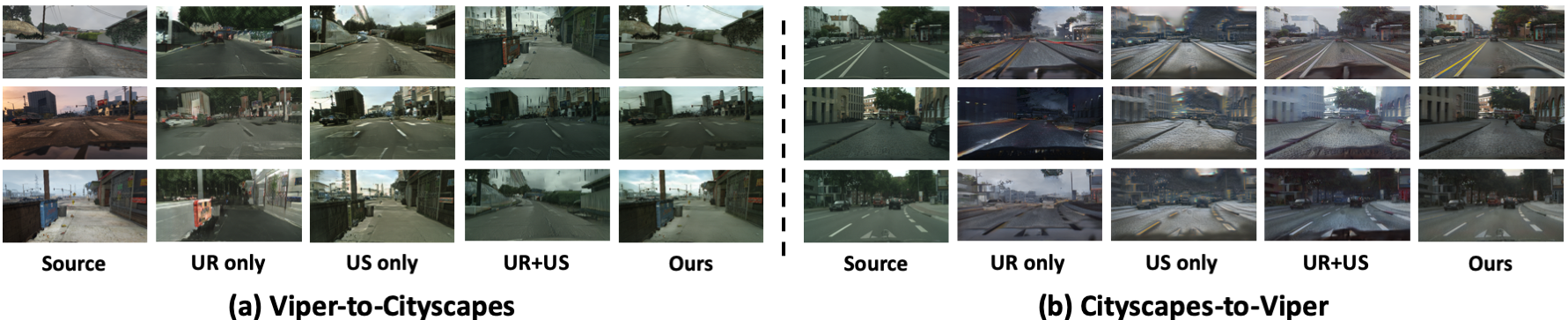}
  \caption{Ablation studies for the Viper-to-Cityscapes translation.}
   \label{fig:ab_v2c}
\end{figure*}

\begin{figure*}[t!]
\centering
   \includegraphics[width=1\linewidth]{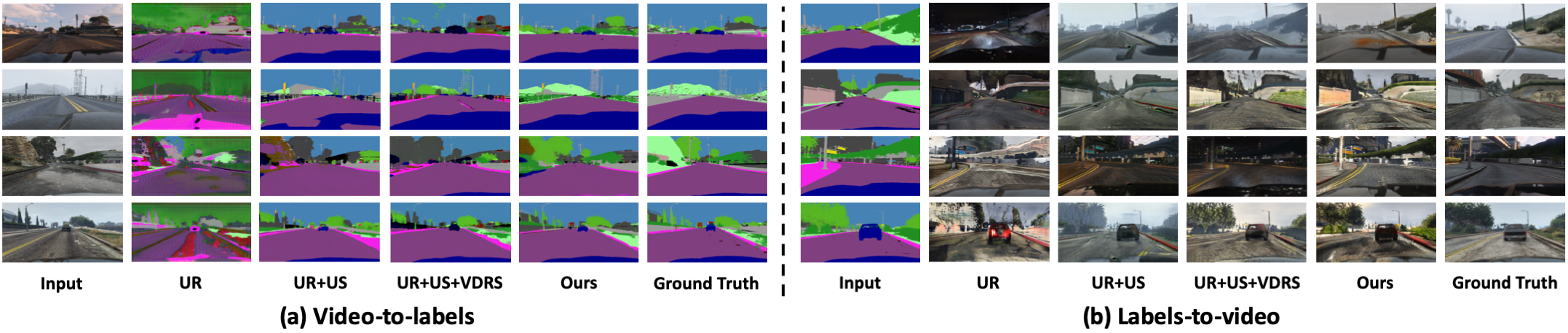}
  \caption{Ablation studies for the video-to-label translation.}
   \label{fig:ab_v2l}
\end{figure*}

\textbf{Ablation studies on synthetic optical flow}: To further explore the effect of the synthetic optical flow $\Tilde{f}$, we conduct ablation studies in both the Viper-to-Cityscapes and video-to-labels translation tasks. We at first investigate the influence of the Gaussian noise $\Delta$ and run a baseline experiment with $\Delta=0$. 
We also verify the importance of the accuracy in the pseudo-supervision brought by the synthetic optical flow. To be more specific, we simulate the process of inaccurate estimation and run an experiment with a different synthetic optical flow for the warping operation in each of the source and target domains. Therefore, the mismatch of the motion in the source and target domains will mislead the training of the generators.
Results of experiments evaluated in warping error, segmentation score, and FCN-score are presented in Table~\ref{table:flow_ab}.
\begin{table}[h]
\centering
\setlength{\tabcolsep}{2pt}
\fontsize{9}{10} \selectfont
\begin{tabular}{ c | c | c c c | c c c} 
 \hline
 \multirow{2}{*}{} & \multirow{2}{*}{Warping} & \multicolumn{3}{c|}{V2L} & \multicolumn{3}{c}{L2V}\\ 
 & & mIoU& AC & MP & mIoU& AC & MP \\
 \hline\hline
  No noise& 0.056462& 0.16 & 3.23 & 4.86 & 4.96 & 9.44 & 32.15 \\
  Wrong flow& 0.052347 & 0.46 & 3.23 & 14.22 & 0.66  & 2.75 & 10.16 \\ 
  \hline
  Ours & 0.029555 & 12.29 & 16.13 &63.20 & 10.08 & 15.73 &47.42 \\
 \hline
\end{tabular}
\caption{Ablation studies of our method on video-to-labels translation.}
\label{table:flow_ab}
\end{table}

We can observe from the results that both of these operations greatly undermine the performance of our method in terms of temporal consistency and semantic consistency.
For the experiment without Gaussian noise, the method fails to perform robust translation due to the lack of regularization from noise and thus suffers from severe overfitting problems. 
On the other hand, the method with the wrong optical flow is misled by the inaccurate guidance originated from the mismatched motions between domains, which leads to a significant drop in the stability of translation. Consequently, the comparison of this experiment with our method reveals the advantage and effectiveness of our spatiotemporal consistency constraints based on synthetic optical flow.

\section{Appendix C}

\begin{table}[t]
\centering
\setlength{\tabcolsep}{1.5pt}
\fontsize{9}{10} \selectfont
\begin{tabular}{ c c c c c c c } 
 \hline
 method  & Day & Sunset & Rain  & Snow & Night & Averge\\ 
 \hline\hline
 Recycle-GAN$^{\dagger}$ & 20.39 & 21.32  & 25.67 &  21.44 &  21.45 & 22.05\\
 UVIT & 17.32 & 16.79 & 19.52 & 18.91 & 19.93 & 18.49\\ 
 \hline
 Improvement (\%) & 15.06 & 21.25 & \textbf{23.96} & 11.80 & 7.09  & 16.15 \\ 
 \hline\hline
 MUNIT& 12.08 & 15.78  & 11.90 &  9.67 &  19.96 & 13.88 \\
 Ours$_{munit}$ & 9.40 & 11.30 & 10.23 & 8.44 & 15.13 & 10.90  \\ 
 \hline
 Improvement (\%) & \textbf{22.18} & \textbf{28.39} & 14.00 & \textbf{12.77} & \textbf{24.20} & \textbf{21.47} \\ 
 \hline
\end{tabular}
\caption{Label to video experiments comparing with UVIT. Recycle-GAN$^{\dagger}$ indicates the improved Recycle-GAN model equipped with the multi-frame based multimodal translators in UVIT.}
\label{table:SOTA_l2v_UVIT}
\end{table}

\textbf{Comparison with UVIT in Labels-to-video translation}:
Since UVIT evaluates the quality of the generated images through Fr\'echet Inception Distance (FID), we also and apply a pretrained network~\cite{CarreiraZ17} as a feature extractor and compute FID following UVIT.
Considering that UVIT adopts a multimodal translation paradigm on multi-frame based translators, we compute the improvement in FID of our method combined with MUNIT over the MUNIT baseline and compare it with the improvement of the UVIT over the Recycle-GAN equipped with multimodal translation paradigm to exclude the influence brought by the architecture of the translators and present the results in Table~\ref{table:SOTA_l2v_UVIT}.
Based on the results presented in Table~\ref{table:SOTA_l2v_UVIT}, our significant improvements in FCN-scores and FID over baselines reveal the effectiveness of our method in synthesizing images. 
Due to the flexibility of our unsupervised losses, the quality of the generated videos is further improved when we deploy them on the basis of MUNIT.

\subsection{Qualitative results}
% \textbf{Viper-to-Cityscapes translation}: In this section, we present the qualitative results of our Viper-to-Cityscapes translation experiment in Figure~\ref{fig:fig3}. With the deployment of our method, we can clearly observe significantly fewer semantic label flipping problems compared with existing methods in both the forward and backward directions, which shows the improvement in the quality of translation.

% \textbf{Video-to-labels translation}: Meanwhile, qualitative results for the video-to-labels translation is also presented in Figure~\ref{fig:v2l}. We can notice from examples in Figure~\ref{fig:v2l}(a) that our method performs more accurate video-to-labels translation under various circumstances, including local details, e.g., the poles in the third row and the trees in the fifth row, and complicated scenes with overlapping contents, e.g., the regions highlighted in the first, second and the fourth row.
% On the other hand, aided by Figure~\ref{fig:v2l}(b), we can more intuitively observe the enhanced quality in our results with more realistic, natural, and detailed scenes, such as roads and sky in the third and fourth rows, and vegetation at the first, second and fifth rows.

% \textbf{Face-to-face translation}: We present more face-to-face translation results for a more comprehensive understanding of the effect of our method over baselines in Figure~\ref{fig:fig_more_face}.
\textbf{Face-to-face translation}: We tested our method on the face-to-face translation tasks, including the translation between Barack Obama and Donald Trump, as well as between John Oliver and Stephen Colbert. In this scenario, given a source domain video, the translated output is expected to keep the original facial expression from the source video but with style in the target domain, without any manual alignment or supervision. Qualitatively results are presented in Figure~\ref{fig:fig_face}. 
Compared with the baselines, our method tackles all these problems and generates more realistic videos with better local details as well as stylistic features in the target domain. In all examples shown in Figure~\ref{fig:fig_face}, our method is able to accurately translate the facial expression from the source video while keeping the characteristics in the target domain without noticeable artifacts.

\textbf{Ablation studies}: We also illustrate exemplar qualitative results of our ablation studies on the Viper-to-Cityscapes experiment in Figure~\ref{fig:ab_v2c} and video-to-labels translation experiment in Figure~\ref{fig:ab_v2l}  respectively.
For the Viper-to-Cityscapes translation example, we can clearly observe the improvement of the translation due to the deployment of our unsupervised spatiotemporal consistency constraints, combining the results presented in Table~\ref{table:ablation_v2c}.
Similar conclusion can be drawn when observing the results in Figure~\ref{fig:ab_v2l} and Table~\ref{table:ablation_v2l}. Moreover, we especially note that the application of our video domain regularization suppression effectively improves the quality of the generated video with enhanced local details in Figure~\ref{fig:ab_v2l}(b).

\end{document}